# Efficient Aircraft Design Optimization Using Multi-Fidelity Models and Multi-fidelity Physics Informed Neural Networks

**Apurba Sarker**

Department of Mechanical Engineering, Bangladesh University of Engineering and Technology, Dhaka 1000, Bangladesh

**Abstract**

Aircraft design optimization traditionally relies on computationally expensive simulation techniques such as Finite Element Method (FEM) and Finite Volume Method (FVM), which, while accurate, can significantly slow down the design iteration process. The challenge lies in reducing the computational complexity while maintaining high accuracy for quick evaluations of multiple design alternatives. This research explores advanced methods, including surrogate models, reduced-order models (ROM), and multi-fidelity machine learning techniques, to achieve more efficient aircraft design evaluations. Specifically, the study investigates the application of Multi-fidelity Physics-Informed Neural Networks (MPINN) and autoencoders for manifold alignment, alongside the potential of Generative Adversarial Networks (GANs) for refining design geometries. Through a proof-of-concept task, the research demonstrates the ability to predict high-fidelity results from low-fidelity simulations, offering a path toward faster and more cost-effective aircraft design iterations.

## 1. Introduction

Developing methods to replace FEM and FVM based simulation has been in the interest of researchers for a long time. FEM and FVM can produce accurate results under the condition of fine meshing but solving fine meshes is time consuming and computationally expensive. In the recent developments of Multidisciplinary Design Optimization (MDO) it is a necessity to produce accurate results in a short time and also being computationally cheap so that multiple design iterations can be tested in a very short time. The existing literatures attempt to solve this problem mainly focus on Surrogate Modeling, Reduced Order Modeling (ROM) and Mult fidelity Machine Learning techniques.

Starting with surrogate-based strategies Rasmussen et al. (2009) used polynomial regression for exploring optimal regions and design trends in joined-wing aircraft [1]. Total weight was predicted from design parameters. Surrogate models have been trained with lift and drag coefficients from rigid Reynolds-averaged Navier–Stokes (RANS) computations [2]. Radial basis function (RBF) models have been used for scalar outputs like lift, drag, weight, and structural failure from high-fidelity analyses [3]. Regression models have been developed for wing structural mass prediction and maximum von Mises stress estimation [4]. Reliability based optimization [5], certification constrains evaluation [6] and uncertainty quantification [7] have also been done using surrogate models.

The biggest limitation of these implementations is that they are predominantly used to predict scalar quantities. For MDO purposes it is very important for designers to know field quantities like location of maximum stress and displacement of a wing. This required effective analysis of properties like stress and displacement for the entire grid of the simulation. Multi-fidelity models are also well explored for scalar surrogate models, but field quantities pose their own challenges. Multi-fidelity approaches require consideration of grid topology and field features that vary over grids.

For prediction of field quantities POD-based ROMs have been proposed for aeroelastic analysis using CFD for the F-16 aircraft [8] and estimation of structural health from sensor data [9]. Subspace projection model reduction has been used to obtain transonic gust loads [10]. ROMs have also been used for measuring aerodynamic influence coefficients based on steady CFD solutions [11] and limit cycle oscillations analysis [12].

Dealing with the grid difference issues three main approaches were adopted in literature. Enforcing same discretization for both the high and low fidelity grids, Mapping the results of both grids on a common grid and the manifold alignment-based ROM (MA-ROM) approach. Enforcing the same discretization has been used for aerodynamic analysis of projectiles [13] and road vehicles [14]. The mapping to a common grid approach have been used for analysis of transonic air foil [15] and compressor blades [16].

MA-ROM performs better for predicting displacement fields compared to von Mises stress fields [17]. This suggests that the MA-ROM method struggles with more complex field data, especially when the data exhibits intricate features that require many modes to accurately represent the solution. The MA-ROM method assumes that the high- and low-fidelity data are interrelated and that a mapping between them can be found. However, if the discrepancy between high- and low-fidelity datasets is too large, the predictive performance may not be superior to single-fidelity models. Again, the success of the MA-ROM method depends on choosing appropriate parameters for the manifold alignment and the selection of the number of POD modes. If these parameters are not chosen correctly, the method may not align the datasets effectively. These issues may become dominant for MDO based tasks as designs are changed and iterated one after another, the correlation of the grids may become complex and different set of parameters may become more effective. Machine Learning techniques like autoencoders and CycleGAN can be used to learn complex relationships between grids. Manifold-aware CycleGAN has been used to align and generate diffusion tensor imaging (DTI) data on the Riemannian manifold, addressing the alignment challenges in manifold-valued data [18]. Neural network-based CycleGAN approach have been used for manifold fitting, enhancing nonlinear data analysis by efficiently modeling and projecting data onto latent manifolds [19]. Improvements of ROM has been suggested by combining non-linear manifold methods with autoencoders and hyper-reduction in 2D dynamical system predictions [20].

Furthermore, Physics informed neural networks (PINNs) are likely to be effective in changing grid problems. PINN has been used to infer 3D velocity and pressure fields from Tomo-BOS temperature data, offering a promising approach for experimental fluid mechanics analysis [21]. PINNs have been used to reconstruct dense velocity fields, infer pressure, and estimate unknown system parameters like viscosity and friction from sparse, noisy velocity data in a turbulent fluid flow [22]. PINNs have also been used in multi-fidelity models to predict static variables. Multi-fidelity physics-informed neural network (MPINN) has been used to improve computational efficiency in CFD simulations, achieving high accuracy [23]. PINNs can learn about the intricate physics in the grid. This is very important for low fidelity grids as physics is the most reliable parameter in this case. The prospective use of Hierarchical Deep Neural Networks (HiDeNNs) are also interesting in this aspect as they can learn both from data and physics using transfer learning [24].

## 2. Methodology

MPINNs can be used to capture the complex relationships in the grid and the field variables. The link between low-fidelity and high-fidelity data is critical for any multi-fidelity modeling. This link was expressed as follows in a prior study [25]:

$$\hat{y}_{HF} = \rho(x) \cdot y_{LF}(x) + \delta(x)$$

Where  and  represent high and low fidelity data, respectively, $\rho(x)$ represents the multiplicative correlation factor, and $\delta(x)$ represents the additive correlation factor. One of this relationship's flaws is that it can only manage linear correlation between two fidelity datasets, but in some other cases [26] the relation between the low and high-grade data is nonlinear. For those cases, the relation between high and low-fidelity data is expressed as

$$\hat{y}_{HF} = F(x, y_{LF})$$

F(.) is an unknown (nonlinear/linear) function that correlates low- to high-fidelity data, which was investigated by Meng and Karniadakis [27]. They divided F(.) into nonlinear and linear components, which are written as follows:

$$F = F_l + F_{nl}$$

where the linear and nonlinear factors in F are  and , respectively. As a result, the relationship between low-fidelity and high-fidelity data is as follows:

$$\hat{y}_{HF} = \alpha F_l(x, y_{LF}) + (1 - \alpha) F_{nl}(x, y_{LF})$$

where $\alpha$ is a hyper-parameter that specifies the degree of nonlinearity between low-fidelity and high-fidelity data, and the MPINN must be trained to determine the linear–nonlinear relationships, as well as the hyper-parameter.

Meng and Karniadakis [27] designed and validated the MPINN architecture. Three fully connected neural networks form the MPINN. The first one (NNL) approximates low-fidelity data while the second (NNH1) and third (NNH2), together denoted as NNH, estimate the linear () and nonlinear () correlations between the low- and high-grade data, respectively. The predictive accuracy of MPINN is strongly influenced by its size (depth and width). As there is so much low-fidelity data, finding the right size for NNL to resemble a low-fidelity function is straightforward. However, few numbers of high-fidelity data are available due to its computational expense. So, importance should be given to the size of NNH while designing MPINN. Meng and Karniadakis [27] proposed the best range for NNH2 with depth (l) and width (w) as l ∈ [1,2] and w ∈ [4,20]. The architecture is illustrated in Fig. 1.

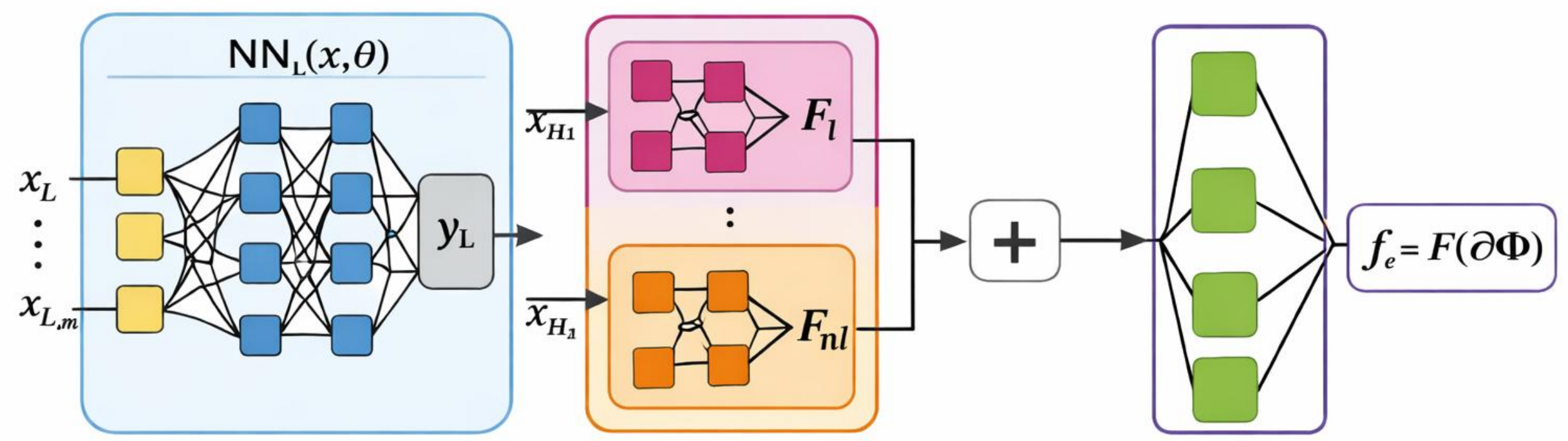

*Fig. 1. Schematic of the multi-fidelity DNN and MPINN. The left box (blue nodes) represents the low-fidelity DNN NNL(x, theta) connected to the box with green dots representing two high-fidelity DNNs, NNHi(x, yL, gamma_i) (i = 1, 2). In the case of MPINN, the combined output of the two high-fidelity DNNs is input to an additional PDE-induced DNN.*

## 3. Implementation on multi-fidelity data

A proof-of-concept task was performed. The task is to predict absolute pressure as a field variable from a low fidelity dataset of a 2D NACA 2412 airfoil under 10m/s air flow. The high and low-fidelity datasets were generated using ANSYS Fluent simulation. The low-fidelity grid had 870 nodes, and the high-fidelity grid had 21,630 nodes. The meshes are shown below.

Fig 2. Multi-fidelity dataset generated using ANSYS Fluent 2D simulation of NACA 2412 airfoil.

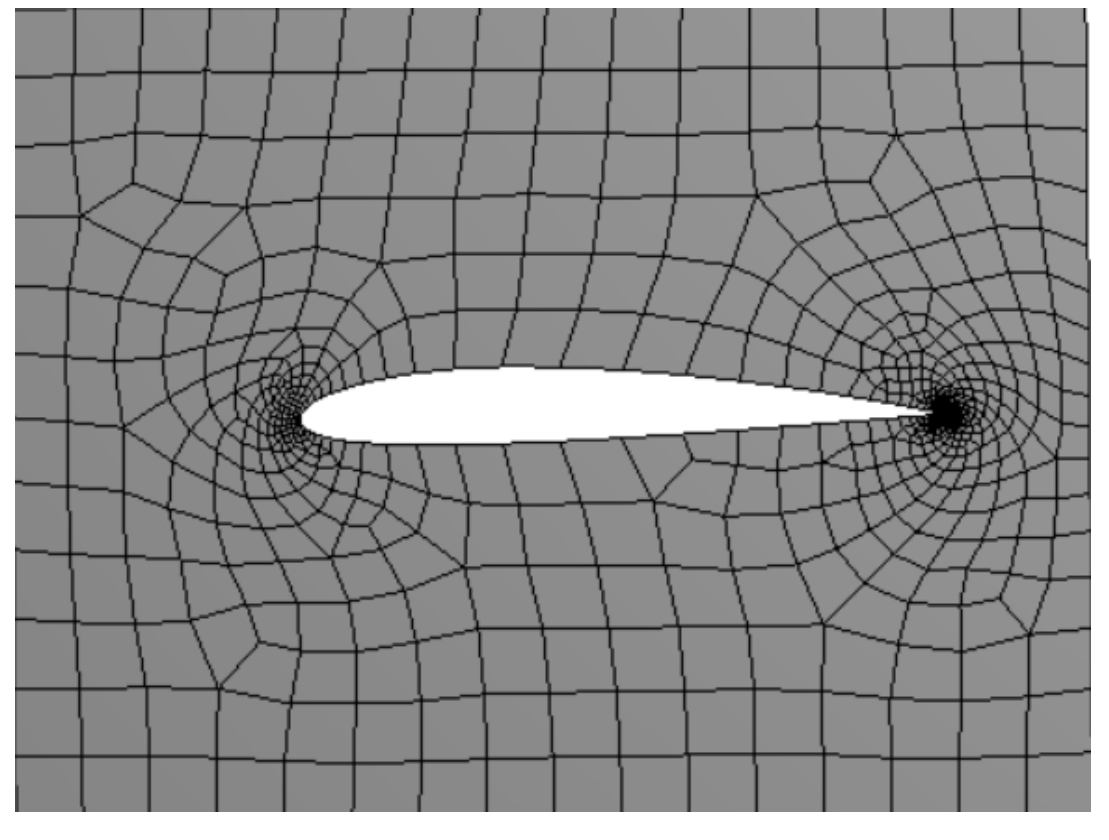

The low-fidelity dataset with 870 nodes. Absolute pressure is calculated in these nodes.

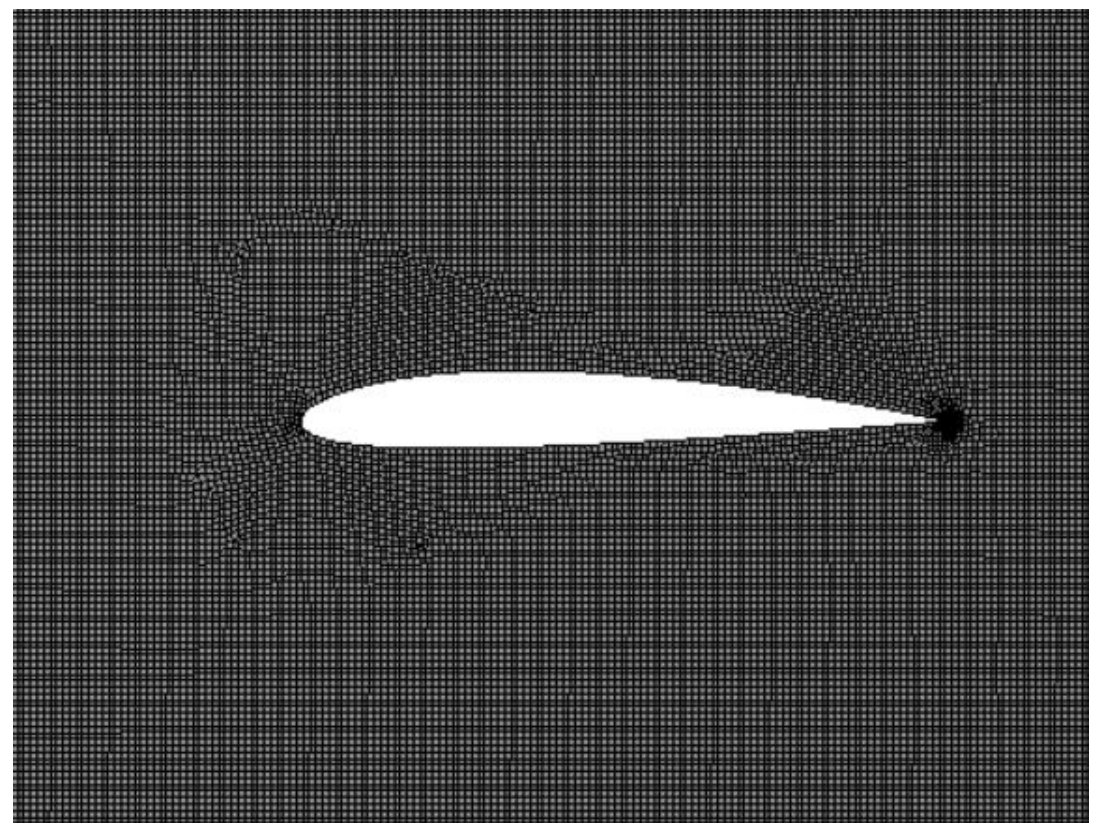

The high-fidelity dataset with 21,630 nodes. Absolute pressure is calculated in these nodes.

*Fig. 2. Multi-fidelity dataset generated using ANSYS Fluent 2D simulation of NACA 2412 airfoil.*

The input dataset for the model is the x and the y co-ordinate and the output is the absolute pressure in the heigh fidelity grid.

The data processing in this implementation involves preparing both the low-fidelity and high-fidelity datasets for training the neural network model. The workflow begins with loading the data from CSV files, which contain the spatial coordinates (x and y) and the corresponding pressure values for both low and high-fidelity simulations.

To align the datasets, interpolation is performed using the griddata method. The high-fidelity pressure values are resampled onto the grid of the low-fidelity dataset. This step ensures consistency between the datasets, allowing the neural network to effectively learn the relationship between low- and high-fidelity data. This step can be improved.

The neural network in this code is designed to predict high-fidelity pressure values from low-fidelity data, leveraging the concept of multi-fidelity learning with physics-informed corrections. The architecture consists of three key components: a non-linear model (NNL) and two correction networks (NNH1 and NNH2).

The non-linear model (NNL) takes the low-fidelity input data, which consists of spatial coordinates, and processes it through three fully connected layers with ReLU activation functions. This network generates the initial low-fidelity pressure predictions, essentially creating a baseline model for the pressure field at lower resolution.

Next, the correction networks (NNH1 and NNH2) aim to refine these low-fidelity predictions using high-fidelity data. NNH1 applies a linear correction to adjust the predictions, while NNH2 introduces a nonlinear correction with L2 regularization, allowing the model to capture more complex relationships between the low- and high-fidelity data. These correction networks are essential for bridging the gap between the low-resolution input and the higher-resolution output.

The full MPINN model integrates the outputs of the three networks. It first predicts low-fidelity pressure values using the NNL, and then adds the corrections from NNH1 and NNH2 to this prediction. The final output of the model is the high-fidelity pressure field, which is learned by minimizing a loss function that includes the Mean Squared Error (MSE) between both the predicted low- and high-fidelity pressures and their respective true values.

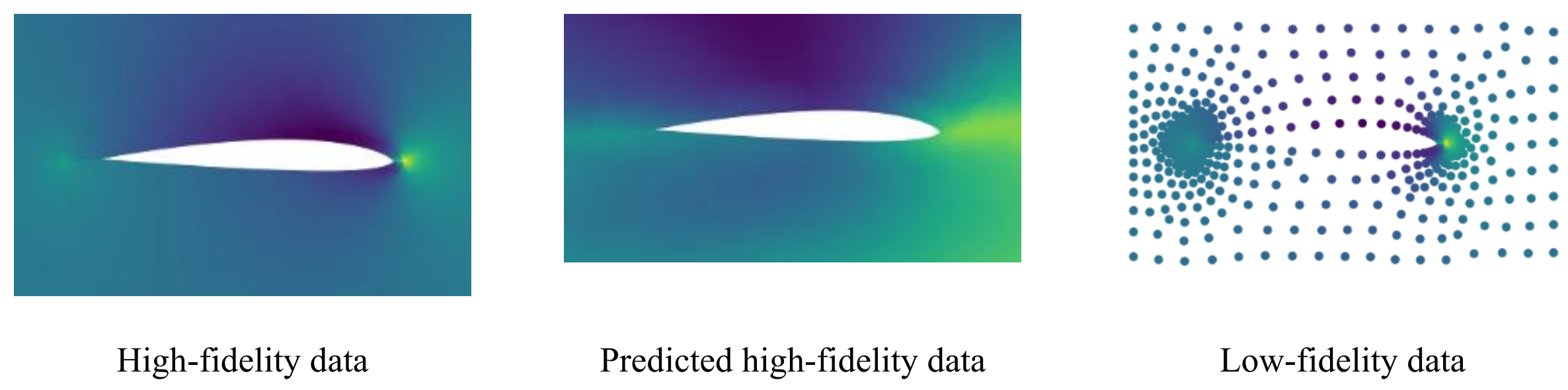

*Fig. 3. Results from the MPINN model.*

## 4. Discussion

The results of this study indicate that advanced multi-fidelity modeling, particularly using MPINN and autoencoders for manifold alignment, can effectively bridge the gap between low-fidelity and high-fidelity datasets. This enables the prediction of high-fidelity outcomes with minimal computational resources, which is essential in optimizing aircraft design processes. The proof-of-concept task demonstrated the feasibility of these methods, confirming that high-fidelity simulations can be approximated with sufficient accuracy, significantly reducing the time and cost typically required for simulations.

Moreover, the integration of machine learning techniques provides a promising pathway to automate the design iteration process. As shown in the model, incorporating corrections from both linear and nonlinear networks allows for greater flexibility in handling complex data relationships. The success of this approach opens the door for more sophisticated models that could handle even more intricate design features and datasets.

The potential for future applications lies in integrating Generative Adversarial Networks (GANs) and Model-Order Reduction (MA-ROM) into the design workflow. These tools will automate the optimization process, continuously refining design geometries based on performance feedback. The iterative nature of this approach promises to accelerate aircraft design, leading to faster turnaround times and more cost-effective solutions in the aerospace industry. However, challenges remain, particularly in optimizing the neural network architectures and ensuring the reliability of predictions across diverse design scenarios.

## 5. Conclusion

In conclusion, the research demonstrates that advanced multi-fidelity modeling approaches, particularly through machine learning techniques such as MPINN and autoencoders for manifold alignment, can significantly improve the efficiency of aircraft design evaluations. These methods allow for the prediction of high-fidelity results from low-fidelity datasets, reducing the need for computationally expensive simulations without sacrificing accuracy.

The success of the proof-of-concept work paves the way for further developments aimed at automating the aircraft design process. By integrating Generative Adversarial Networks (GANs) and Model-Order Reduction (MA-ROM) techniques into the design workflow, future efforts will create a fully automated, optimized design process that accelerates decision-making and reduces computational costs, driving the future of aircraft design toward greater efficiency.

## 6. References

[1] C.C. Rasmussen, R.A. Canfield, M. Blair, Optimization process for configuration of flexible joined-wing, Struct Multidisc Optim 37 (2009) 265–277. https://doi.org/10.1007/s00158-008-0229-4.

[2] M. Bordogna, D. Bettebghor, C. Blondeau, R. De Breuker, Surrogate-based aerodynamics for composite wing box sizing., 17th International Forum Aeroelasticity Structure Dynamics (n.d.).

[3] A. Benaouali, S. Kachel, Multidisciplinary design optimization of aircraft wing using commercial software integration, Aerospace Science and Technology 92 (2019) 766–776. https://doi.org/10.1016/j.ast.2019.06.040.

[4] V. Cipolla, K. Abu Salem, G. Palaia, V. Binante, D. Zanetti, A DoE-based approach for the implementation of structural surrogate models in the early stage design of box-wing aircraft, Aerospace Science and Technology 117 (2021) 106968. https://doi.org/10.1016/j.ast.2021.106968.

[5] X. Li, C. Gong, L. Gu, Z. Jing, H. Fang, R. Gao, A reliability-based optimization method using sequential surrogate model and Monte Carlo simulation, Struct Multidisc Optim 59 (2019) 439–460. https://doi.org/10.1007/s00158-018-2075-3.

[6] D. Sarojini, J. Xie, Y. Cai, J.A. Corman, D. Mavris, A Certification-Driven Platform for Multidisciplinary Design Space Exploration in Airframe Early Preliminary Design, in: AIAA AVIATION 2020 FORUM, American Institute of Aeronautics and Astronautics, VIRTUAL EVENT, 2020. https://doi.org/10.2514/6.2020-3157.

[7] R. Duca, D. Sarojini, S. Bloemer, I. Chakraborty, S.I. Briceno, D.N. Mavris, Effects of Epistemic Uncertainty on Empennage Loads During Dynamic Maneuvers, in: 2018 AIAA Aerospace Sciences Meeting, American Institute of Aeronautics and Astronautics, Kissimmee, Florida, 2018. https://doi.org/10.2514/6.2018-0767.

[8] T. Lieu, C. Farhat, Adaptation of Aeroelastic Reduced-Order Models and Application to an F-16 Configuration, AIAA Journal 45 (2007) 1244–1257. https://doi.org/10.2514/1.24512.

[9] L. Mainini, K. Willcox, Surrogate Modeling Approach to Support Real-Time Structural Assessment and Decision Making, AIAA Journal 53 (2015) 1612–1626. https://doi.org/10.2514/1.J053464.

[10] P. Bekemeyer, S. Timme, Flexible aircraft gust encounter simulation using subspace projection model reduction, Aerospace Science and Technology 86 (2019) 805–817. https://doi.org/10.1016/j.ast.2019.02.011.

[11] M. Ripepi, M.J. Verveld, N.W. Karcher, T. Franz, M. Abu-Zurayk, S. Görtz, T.M. Kier, Reduced-order models for aerodynamic applications, loads and MDO, CEAS Aeronaut J 9 (2018) 171–193. https://doi.org/10.1007/s13272-018-0283-6.

[12] K. Lu, Y. Jin, Y. Chen, Y. Yang, L. Hou, Z. Zhang, Z. Li, C. Fu, Review for order reduction based on proper orthogonal decomposition and outlooks of applications in mechanical systems, Mechanical Systems and Signal Processing 123 (2019) 264–297. https://doi.org/10.1016/j.ymssp.2019.01.018.

[13] M.J. Mifsud, D.G. MacManus, S.T. Shaw, A variable-fidelity aerodynamic model using proper orthogonal decomposition, Numerical Methods in Fluids 82 (2016) 646–663. https://doi.org/10.1002/fld.4234.

[14] A. Bertram, C. Othmer, R. Zimmermann, Towards Real-time Vehicle Aerodynamic Design via Multi-fidelity Data-driven Reduced Order Modeling, in: 2018 AIAA/ASCE/AHS/ASC Structures, Structural Dynamics, and Materials Conference, American Institute of Aeronautics and Astronautics, Kissimmee, Florida, 2018. https://doi.org/10.2514/6.2018-0916.

[15] B. Malouin, J.-Y. Trépanier, M. Gariépy, Interpolation of Transonic Flows Using a Proper Orthogonal Decomposition Method, International Journal of Aerospace Engineering 2013 (2013) 1–11. https://doi.org/10.1155/2013/928904.

[16] T. Benamara, P. Breitkopf, I. Lepot, C. Sainvitu, P. Villon, Multi-fidelity POD surrogate-assisted optimization: Concept and aero-design study, Struct Multidisc Optim 56 (2017) 1387–1412. https://doi.org/10.1007/s00158-017-1730-4.

[17] C. Perron, D. Sarojini, D. Rajaram, J. Corman, D. Mavris, Manifold alignment-based multi-fidelity reduced-order modeling applied to structural analysis, Struct Multidisc Optim 65 (2022) 236. https://doi.org/10.1007/s00158-022-03274-1.

[18] B. Anctil-Robitaille, C. Desrosiers, H. Lombaert, Manifold-Aware CycleGAN for High-Resolution Structural-to-DTI Synthesis, in: N. Gyori, J. Hutter, V. Nath, M. Palombo, M. Pizzolato, F. Zhang (Eds.), Computational Diffusion MRI, Springer International Publishing, Cham, 2021: pp. 213–224. https://doi.org/10.1007/978-3-030-73018-5_17.

[19] Z. Yao, J. Su, S.-T. Yau, Manifold fitting with CycleGAN, Proc. Natl. Acad. Sci. U.S.A. 121 (2024) e2311436121. https://doi.org/10.1073/pnas.2311436121.

[20] F. Romor, G. Stabile, G. Rozza, Non-linear manifold ROM with Convolutional Autoencoders and Reduced Over-Collocation method, (2022). https://doi.org/10.48550/ARXIV.2203.00360.

[21] S. Cai, Z. Wang, F. Fuest, Y.J. Jeon, C. Gray, G.E. Karniadakis, Flow over an espresso cup: inferring 3-D velocity and pressure fields from tomographic background oriented Schlieren via physics-informed neural networks, J. Fluid Mech. 915 (2021) A102. https://doi.org/10.1017/jfm.2021.135.

[22] V. Parfenyev, M. Blumenau, I. Nikitin, Inferring Parameters and Reconstruction of Two-Dimensional Turbulent Flows with Physics-Informed Neural Networks, Jetp Lett. 120 (2024) 599–607. https://doi.org/10.1134/S0021364024602203.

[23] M.D. Rony, M. Islam, Md.A. Islam, M.N. Hasan, Fast Predictive Artificial Neural Network Model Based on Multi-fidelity Sampling of Computational Fluid Dynamics Simulation, in: Md.S. Hossain, S.P. Majumder, N. Siddique, Md.S. Hossain (Eds.), The Fourth Industrial Revolution and Beyond, Springer Nature Singapore, Singapore, 2023: pp. 103–116. https://doi.org/10.1007/978-981-19-8032-9_8.

[24] S. Saha, Z. Gan, L. Cheng, J. Gao, O.L. Kafka, X. Xie, H. Li, M. Tajdari, H.A. Kim, W.K. Liu, Hierarchical Deep Learning Neural Network (HiDeNN): An artificial intelligence (AI) framework for computational science and engineering, Computer Methods in Applied Mechanics and Engineering 373 (2021) 113452. https://doi.org/10.1016/j.cma.2020.113452.

[25] M. Giselle Fernández-Godino, Lawrence Livermore National Laboratory, Livermore, California 94550, USA, Review of multi-fidelity models, ACSE 1 (2023) 351–400. https://doi.org/10.3934/acse.2023015.

[26] H. Babaee, P. Perdikaris, C. Chryssostomidis, G.E. Karniadakis, Multi-fidelity modelling of mixed convection based on experimental correlations and numerical simulations, J. Fluid Mech. 809 (2016) 895–917. https://doi.org/10.1017/jfm.2016.718.

[27] X. Meng, G.E. Karniadakis, A composite neural network that learns from multi-fidelity data: Application to function approximation and inverse PDE problems, Journal of Computational Physics 401 (2020) 109020. https://doi.org/10.1016/j.jcp.2019.109020.

[28] J. Chattoraj, J.C. Wong, Z. Zexuan, M. Dai, X. Yingzhi, L. Jichao, X. Xinxing, O.C. Chun, Y. Feng, D.M. Ha, L. Yong, Tailoring Generative Adversarial Networks for Smooth Airfoil Design, (2024). https://doi.org/10.48550/ARXIV.2404.11816.